\documentclass[journal]{IEEEtran}
\usepackage{slashbox,graphicx,times,amsmath,amssymb,cite,subfigure,stfloats,booktabs, url,multirow, footnote,warpcol,graphicx,dcolumn,enumitem}
\usepackage[flushleft]{threeparttable}

\begin{document}
\title{Canonical Correlation Analysis (CCA) Based Multi-View Classification: An Overview}
\author{
Chenfeng~Guo and Dongrui~Wu
\thanks{C.~Guo and D.~Wu are with the Key Laboratory of the Ministry of Education for Image Processing and Intelligent Control, School of Artificial Intelligence and Automation, Huazhong University of Science and Technology, Wuhan 430074, China. Email: cfguo@hust.edu.cn, drwu@hust.edu.cn.
    }
    \thanks{
        Dongrui~Wu is the corresponding author.
    }
}
\maketitle

\begin{abstract}
Multi-view learning (MVL) is a strategy for fusing data from different sources or subsets. Canonical correlation analysis (CCA) is very important in MVL, whose main idea is to map data from different views onto a common space with maximum correlation. Traditional CCA can only be used to calculate the linear correlation of two views. Besides, it is unsupervised and the label information is wasted. Many nonlinear, supervised, or generalized extensions have been proposed to overcome these limitations. However, to our knowledge, there is no overview for these approaches. This paper provides an overview of many representative CCA-based MVL approaches.
\end{abstract}

\begin{IEEEkeywords}
Canonical correlation analysis, multi-view learning
\end{IEEEkeywords}

\IEEEpeerreviewmaketitle

\section{Introduction} \label{sect:Intro}

Many real-world datasets can be described from multiple ``viewpoints", such as pictures taken from different angles of the same object, different language expressions of the same semantic, texts and images on the same web page, etc. The representations from different perspectives can be treated as different views. The essence of multi-view learning (MVL) is to exploit the consensual and complementary information between different views \cite{Tang2018,Xu2013} to achieve better learning performance.

MVL approaches can be divided into three major categories \cite{Xue2017,Baltrusaitis2019}:
\begin{enumerate}
\item \emph{Co-training} \cite{Blum1998,Abney2002}, which exchanges discriminative information between two views by training the two models alternately.
\item \emph{Multi-kernel learning} \cite{Gehler2009,Gonen2011}, which maps data to different feature spaces with different kernels, and then combines those projected features from all spaces.
\item \emph{Subspace learning} \cite{Hotelling1936,Diethe2008,Xia2010}, which assumes all views are generated from a latent common space where shared information of all views can be exploited.
\end{enumerate}

\begin{figure}[htpb] \centering
\includegraphics[width=.98\linewidth,clip]{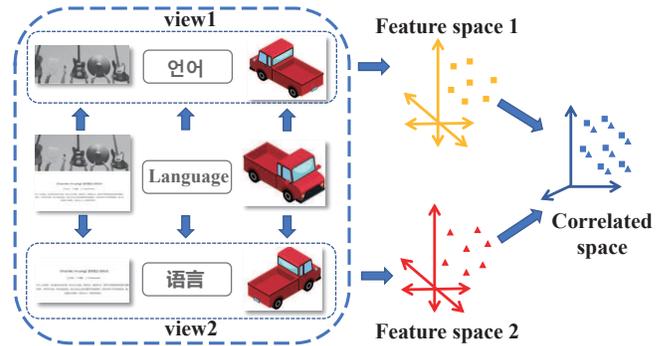}
\caption{Multi-view data and CCA-based subspace learning.} \label{fig:CCA}
\end{figure}

Canonical correlation analysis (CCA), first proposed by Hotelling \cite{Hotelling1936} in 1936, is a typical subspace learning approach. Its main idea is to find pairs of projections for different views so that the correlations between them are maximized. One example is illustrated in Fig.~\ref{fig:CCA}. Since CCA takes the relationship between different feature sets into account, which is consistent with the idea of MVL, it is widely used in MVL \cite{Zhao2017,Sun2013}, including multi-view dimensionality reduction \cite{Foster2008}, multi-view clustering \cite{Kamalika2009,Blaschko2008}, multi-view regression \cite{Kakade2007}, and so on.

Traditional CCA has the following limitations:
\begin{enumerate}
\item It cannot handle more than two views.
\item It can only calculate the linear correlation between two views, whereas in many real-world applications the true relationship between the views may be nonlinear.
\item In supervised classification, labels are available; however, CCA, as an unsupervised algorithm, completely ignores the labels, and hence wastes information.
\end{enumerate}
Many extensions have been proposed in the past few decades \cite{Akaho2006,Sun2007,Wang2015DCCAE,Sun2008,Andrew2013,Benton2016,Elmadany2016,Luo2015,Nielsen2002} to accommodate these limitations. Some representative ones are briefly introduced next.

In 1961, Horst \cite{Horst1961} first proposed generalized canonical correlation analysis (GCCA) to estimate the pairwise correlations of multiple views. He provided two formulations: the sum of correlation (SUMCOR), and the maximum variance (MAXVAR). Carroll \cite{Carroll1968} in 1968 proposed to find a shared latent correlated space, which was shown to be identical to MAXVAR GCCA. In 1971, Kettenring \cite{Kettenring1971} added three new formulations to GCCA, which maximizes the sum of the squared correlation (SSQCOR), minimizes the smallest eigenvalue (MINVAR), and minimizes the determinant of the correlation matrix (GENVAR), respectively. Two decades later, Nielsen \cite{Nielsen2002} summarized four constraints for these five formulations, forming a total of 20 combinations. In 2007, Via \cite{Via2007} proposed least squares based CCA (LS-CCA), and showed that it is essentially identical to MAXVAR. Luo \emph{et al.} \cite{Luo2015} proposed in 2015 that the correlation of multiple views can be directly maximized by analyzing the high-order covariance tensor, namely tensor canonical correlation analysis (TCCA). In 2017, Benton \emph{et al.} \cite{Benton2016} proposed DNN-based deep generalized canonical correlation analysis (DGCCA), which was a nonlinear extension of MAXVAR GCCA.

Generally, there are three approaches to deal with complex nonlinear relationship between two views \cite{Sun2007}. The most common one is to project data onto a higher dimensional space using the ``kernel trick", e.g., kernel canonical correlation analysis (KCCA) \cite{Akaho2006}. However, global kernelization suffers from high computational complexity, and it is not easy to choose an optimal kernel function. The second approach tries to preserve the locality of data. Inspired by the graph model, Sun and Chen  \cite{Sun2007} proposed locality preserving canonical correlation analysis (LPCCA), which aimed to reduce the global nonlinear dimensionality while preserving the local linear structure of data. Since the distances between neighbors need to be calculated, it is time-consuming when the sample size is large. The third approach is based on deep neural networks (DNN), which can give a very complex mapping between data. Andrew \emph{et al.} \cite{Andrew2013} first proposed deep canonical correlation analysis (DCCA) in 2013. Inspired by the autoencoder, Wang \emph{et al.} \cite{Wang2015DCCAE} proposed deep canonically correlated autoencoders (DCCAE) in 2015. However, DNN models have poor interpretability, and require a large amount of data to fit.

The label information is critical for classification problems. In order to make full use of the discriminant information, Sun \emph{et al.} \cite{Sun2008} proposed discriminant canonical correlation analysis (DisCCA) in 2008, by taking the inter-class and intra-class similarities of different views into consideration. Similar to DisCCA, Sun \emph{et al.} \cite{Sun2016} further proposed multi-view linear discriminant analysis (MLDA), which combined CCA and linear discriminant analysis (LDA) \cite{Fisher1936}. Elmadany \emph{et al.} \cite{Elmadany2016} integrated neural networks into DisCCA to obtain a nonlinear supervised model. Benton \emph{et al.} \cite{Benton2016} made use of the discriminant information by treating the one-hot encoding matrix of the labels as an additional view.

This paper provides an overview of many representative CCA approaches. A summary of their main characteristics is shown in Table~\ref{tab:comparation}.

\begin{table}[htpb] \centering 
\caption{Comparison of some typical CCA-based approaches.} \label{tab:comparation}
\begin{tabular}{l| c c c c} \toprule[0.08em]
\textbf{Approaches}&\textbf{Multiple views}&\textbf{Supervised}&\textbf{Nonlinear}&\textbf{DNN} \\ \hline
    CCA         &  &  &   &    \\
    sCCA  &  &  &   &    \\
    KCCA        &  &  & \checkmark  &    \\
    RCCA        &  &  & \checkmark  &    \\
    LPCCA       &  &  & \checkmark  &    \\
    DisDCCA    &  & \checkmark &   &    \\
    MLDA        &  & \checkmark &   &    \\
    MULDA       &  & \checkmark &   &    \\
    DCCA        &  &  &  \checkmark & \checkmark   \\
    DisDCCA    &  & \checkmark & \checkmark  &  \checkmark  \\
    DCCAE       &  &  & \checkmark  & \checkmark   \\
    DisDCCAE   &  & \checkmark & \checkmark  &   \checkmark   \\
    VCCA        &  &  & \checkmark  &  \checkmark   \\
    GCCA        & \checkmark &  &   &    \\
    MCCA        & \checkmark &  &   &    \\
    LS-CCA      & \checkmark &  &   &    \\
    TCCA        & \checkmark &  &   &    \\
    DGCCA       & \checkmark & \checkmark &  \checkmark &   \checkmark   \\
\bottomrule
\end{tabular}
\end{table}

The remainder of this paper is organized as follows: Sections~\ref{sec:TwoViewCCA} and \ref{sec:MultiViewCCA}  introduce CCA-based MVL approaches for two views and more than two views, respectively. Section~\ref{sec:application} provides their applications. Section~\ref{sec:Conclusion} draws conclusions.

\section{CCA-Based MVL Approaches for Two Views} \label{sec:TwoViewCCA}

This section first reviews the tradition CCA, and then introduces some representative nonlinear, sparse, and/or supervised extensions. Table~\ref{tab:notation} summarizes the notations used in this paper.

\begin{table}[htpb] \centering 
\caption{List of notations.} \label{tab:notation}
\begin{tabular}{l l}
\toprule[0.08em]
\textbf{Notation}&\textbf{Description} \\ \hline
    $X$, $Y$ & Data matrices from two different views $x$ and $y$ \\
    $\left(\mathbf{x}_{i},\mathbf{y}_{i}\right)$ & The $i$-th paired instances from Views $x$ and $y$ \\
    $N$ &	Number of instances in each view\\
    $c$ &   Number of classes \\
    $n_{i}$& Number of instances in Class $i$ \\
    $J$ &   Number of views\\
    $K$ &   Number of canonical vectors \\
    $d_x$ & Feature dimensionality of View $x$ \\
    $\mathbf{w}_x$ &	A canonical vector of View $x$ \\
    $W_x$ &	Canonical matrix of View $x$ \\
    $\Sigma_{xy}$ & Covariance matrix of View $x$ and View $y$\\
    $\hat{\Sigma}_{xy}$ & Regularized covariance matrix of View $x$ and View $y$\\
    $r_x$ & Regularization coefficients of View $x$ \\
    $m$ & Dimension of low rank approximation  \\
    $\cdot^{T}$ & Transpose of a matrix or vector \\
    $I$ & The identity matrix\\
    $\|\cdot\|_{2}$ & 2-norm of a vector \\
    $\|\cdot\|_{F}$ & The Frobenius-norm of a matrix\\
\bottomrule
\end{tabular}
\end{table}

\subsection{Canonical Correlation Analysis (CCA) }\label{subsection:CCA}

Let $X=\left[\mathbf{x}_{1}, \mathbf{x}_{2}, \dots, \mathbf{x}_{N}\right] \in \mathbb{R}^{d_{x} \times N}$ and $Y =\left[\mathbf{y}_{1}, \mathbf{y}_{2}, \dots, \mathbf{y}_{N}\right]\in \mathbb{R}^{d_{y} \times N}$ be two mean-zero data matrices with $N$ instances and $d_x$ and $d_y$ features, respectively. CCA aims to find $K$ pairs of linear projections ${W}_{x}=\left[\mathbf{w}_{x,1}, \mathbf{w}_{x,2}, \dots, \mathbf{w}_{x,K}\right]\in \mathbb{R}^{d_{x}\times K}$ and ${W}_{y}=\left[\mathbf{w}_{y,1}, \mathbf{w}_{y,2}, \dots, \mathbf{w}_{y,K}\right]\in \mathbb{R}^{d_{x}\times K}$, called canonical vectors, so that the correlations between ${W}_{x}^{T}X$ and ${W}_{y}^{T}Y$ are maximized.

Take a canonical vector $\mathbf{w}_{x}\in \mathbb{R}^{d_{x}\times 1}$ for $X$ and a canonical vector $\mathbf{w}_{y}\in \mathbb{R}^{d_{y}\times 1}$ for $Y$ for example. CCA maximizes the correlation coefficient $\rho$ between $X^{T}\mathbf{w}_{x}$ and $Y^{T}\mathbf{w}_{y}$ \cite{Hardoon2004}, i.e.,
\begin{align}
\rho\left(X^{T}\mathbf{w}_{x}, Y^{T}\mathbf{w}_{y}\right)=\frac{\mathbf{w}_{x}^{T}X Y^{T} \mathbf{w}_{y}}{\sqrt{\left(\mathbf{w}_{x}^{T}XX^{T} \mathbf{w}_{x}\right)\left(\mathbf{w}_{y}^{T} Y Y^{T}\mathbf{w}_{y}\right)}}. \label{equ:CCA}
\end{align}

Since (\ref{equ:CCA}) is invariant to the scaling of $\mathbf{w}_{x}$ and $\mathbf{w}_{y}$, it can be transformed into the following constrained form:
\begin{align}
\max _{\mathbf{w}_{x}, \mathbf{w}_{y}}\ & \mathbf{w}_{x}^{T} XY^{T} \mathbf{w}_{y}\\ \nonumber
s.t.\ & \mathbf{w}_{x}^{T}X X^{T} \mathbf{w}_{x}=1, \mathbf{w}_{y}^{T}YY^{T} \mathbf{w}_{y}=1.
\end{align}

When the feature dimensionality is high, especially when $d_x>N$ (or $d_y>N$), the covariance matrix $XX^T$ (or $YY^T$) is singular, and hence the optimization problem is under-determined. Regularizations can be added to the covariance matrices to remedy this problem \cite{Bickel2008,Tijl2003,Warton2008}, by introducing:
\begin{align}
\hat{\Sigma}_{xx}&=\frac{1}{N} X X^{T}+ r_{x} I,\\
\hat{\Sigma}_{yy}&=\frac{1}{N} Y Y^{T}+r_{y} I,
\end{align}
where $r_{x}$ and $r_{y}$ are non-negative regularization coefficients.

There are two approaches for computing $W_x$ and $W_y$ directly. The first is to solve the following generalized eigenvalue decomposition problem \cite{Hardoon2004}:
\begin{align}
\left[ \begin{array}{cc}{\mathbf{0}} & {{\Sigma}_{xy}} \\ {{\Sigma}_{yx}} & {\mathbf{0}}\end{array}\right] \left[ \begin{array}{c}{\mathbf{w}_{\mathbf{x}}} \\ {\mathbf{w}_{\mathbf{y}}}\end{array}\right]=\lambda \left[ \begin{array}{cc}{\hat{\Sigma}_{xx}} & {\mathbf{0}} \\ {\mathbf{0}} & {\hat{\Sigma}_{yy}}\end{array}\right] \left[ \begin{array}{c}{\mathbf{w}_{\mathbf{x}}} \\ {\mathbf{w}_{\mathbf{y}}}\end{array}\right],
\end{align}
where
\begin{align}
{\Sigma}_{xy}&=\frac{1}{N} X Y^{T},\\
{\Sigma}_{yx}&=\frac{1}{N} Y X^{T}.
\end{align}
$\{[\mathbf{w}_{x,k};\mathbf{w}_{y,k}]\}_{k=1}^K$ are then the $K$ leading generalized eigenvectors. The correlation $\rho\left(\mathbf{w}_{x,k}^{T}X, \mathbf{w}_{y,k}^{T}Y\right)$ is equal to the $k$-th leading  generalized eigenvalue.

The second solution \cite{Andrew2013} performs singular value decomposition (SVD) on matrix $T=\hat{\Sigma}_{xx}^{-1 / 2} {\Sigma}_{xy} \hat{\Sigma}_{yy}^{-1 / 2}$. Let $\tilde{W}_{x}$ and $\tilde{W}_{y}$ be the $K$ leading left and right singular vectors of $T$. Then, the canonical matrices are $W_{x}= \hat{\Sigma}_{xx}^{-1 / 2} \tilde{W}_{x}$ and $W_{y}= \hat{\Sigma}_{yy}^{-1 / 2} \tilde{W}_{y}$. The correlation $\rho\left(\mathbf{w}_{x,k}^{T}X, \mathbf{w}_{y,k}^{T}Y\right)$ is equal to the $k$-th leading  singular value of $T$.

Once $W_x$ and $W_y$ are obtained, the projected new features, called canonical variables, are computed by
$Z_{x}=W_{x}^{T} X$ and $Z_{y}=W_{y}^{T} Y$.

\subsection{Sparse CCA (sCCA)}

Canonical matrices $W_{x}$ and $W_{y}$ calculated by CCA is dense. Based on penalized matrix decomposition (PMD) \cite{Witten2009}, Witten \emph{et al.}  \cite{Witten2009Extensions} proposed penalized CCA, called sCCA in this paper, to obtain spare canonical vectors. sCCA also remedies the problem that canonical vectors are not unique when $N$ is smaller than $d_{x}$ and/or $d_{y}$.

PMD \cite{Eckart1936} performs rank-$m$ approximation of an arbitrary matrix $M \in \mathbb{R}^{d \times N}$ by ${\hat{M}}=\sum_{i=1}^{m} d_{i} \mathbf{u}_{i} \mathbf{v}_{i}^{T}$, using SVD, where $d_{i}$ is the $i$-th leading singular value, and $\mathbf{u}_{i}\in \mathbb{R}^{d\times 1}$ and $\mathbf{v}_{i}\in \mathbb{R}^{N\times 1}$ are the corresponding left and right singular vectors, respectively. The sparsity of $\hat{M}$ can be achieved by imposing LASSO-constraints on $\mathbf{u}_{i}$ and $\mathbf{v}_{i}$. When $m=1$, minimizing $\|M-\hat{M}\|_{F}^{2}$ is equivalent to maximizing the following objective function:
\begin{align}
\max_ {\mathbf{u}, \mathbf{v}}\ & \mathbf{u}^{T} M \mathbf{v} \label{eq:PMD}\\
s.t.\ &\|\mathbf{u}\|_{2}^{2}=1,\ \|\mathbf{v}\|_{2}^{2}=1,\ p_{1}(\mathbf{u}) \leqslant c_{1},\ p_{2}(\mathbf{v}) \leqslant c_{2}, \nonumber
\end{align}
where $p_{1}(\mathbf{u})=\sum_{i=1}^d\left|u_{i}\right|$ and $p_2(\mathbf{v})=\sum_{i=1}^N\left|v_{i}\right|$ represent the LASSO penalties, and parameters $c_{1}$ and $c_{2}$ control the degree of sparsity.

sCCA can be solved by replacing $M$ in (\ref{eq:PMD}) with the cross covariance matrix $XY^T$, i.e.,
\begin{align}
\max_ {\mathbf{w}_{x}, \mathbf{w}_{y}}\ & \mathbf{w}_{x}^{T} X Y^T \mathbf{w}_{y}\\ \nonumber
s.t.\ & \|\mathbf{w}_{x}\|_{2}^{2}=1, \|\mathbf{w}_{y}\|_{2}^{2}=1, p_{1}(\mathbf{w}_{x}) \leqslant c_{1}, p_{2}(\mathbf{w}_{y}) \leqslant c_{2}.
\end{align}

$K$ pairs of canonical vectors, $\{\left(\mathbf{w}_{x,k},\mathbf{w}_{x,k}\right)\}_{k = 1}^K$, can be obtained by solving a multi-factor PMD \cite{Witten2009}.

\subsection{Kernel CCA (KCCA)}

The traditional CCA cannot be applied when the correlation between different views is nonlinear. Kernel CCA (KCCA) uses (nonlinear) kernels to project data onto a higher dimensional space for correlation analysis.

Let the projections be $\phi_{x}$ and $\phi_{y}$, and the projected views in the high-dimensional space be $\Phi_{x}=\left[\phi_{x}\left(\mathbf{x}_{1}\right), \phi_{x}\left(\mathbf{x}_{2}\right), \ldots, \phi_{x}\left(\mathbf{x}_{N}\right)\right]$ and $\Phi_{\mathrm{y}}=\left[\phi_{y}\left(\mathbf{y}_{1}\right), \phi_{y}\left(\mathbf{y}_{2}\right), \ldots, \phi_{y}\left(\mathbf{y}_{N}\right)\right]$, respectively. The objective function of KCCA is:
\begin{align}
\max _{\mathbf{w}_{x}, \mathbf{w}_{y}}\ & \mathbf{w}_{x}^{T} \Phi_{x}\Phi_{y}^{T} \mathbf{w}_{y} \label{eq:KCCA} \\
s.t.\ &\mathbf{w}_{x}^{T} \Phi_{x} \Phi_{x}^{T} \mathbf{w}_{x}=1, \mathbf{w}_{y}^{T} \Phi_{y} \Phi_{y}^{T} \mathbf{w}_{y}=1.\nonumber
\end{align}

Expressing $\mathbf{w}_{x}$ and $\mathbf{w}_{y}$ as linear combinations of the columns of ${\Phi}_{x}$ and ${\Phi}_{y}$, respectively \cite{Akaho2006}:
\begin{align}
\mathbf{w}_{x}&=\sum_{i=1}^{N} a^{i} \phi_{x}\left(\mathbf{x}_{i}\right)={\Phi}_{x} \mathbf{a},\\
\mathbf{w}_{y}&=\sum_{i=1}^{N} b^{i} \phi_{y}\left(\mathbf{y}_{i}\right)={\Phi}_{y} \mathbf{b},
\end{align}
where $\mathbf{a}=\left[a^{1},\cdots,a^{N}\right]^{T}$ and $\mathbf{b}=\left[b^{1},\cdots,b^{N}\right]^{T}$ are linear coefficients. (\ref{eq:KCCA}) can then be rewritten as:
\begin{align}
&\max _{\mathbf{a}, \mathbf{b}}\ \mathbf{a}^{T} {\Phi}_{x}^{T} {\Phi}_{x} {\Phi}_{y}^{T} {\Phi}_{y} \mathbf{b} \label{eq:KCCA2}\\
&s.t.\ \mathbf{a}^{T} {\Phi}_{x}^{T} {\Phi}_{x} {\Phi}_{x}^{T} {\Phi}_{x} \mathbf{a}=1,\
\mathbf{b}^{T} {\Phi}_{y}^{T} {\Phi}_{y} {\Phi}_{y}^{T} {\Phi}_{y} \mathbf{b}=1.\nonumber
\end{align}

Let $K_{x}={\Phi}_{x}^{T} {\Phi}_{x}$ and $K_y={\Phi}_{y}^{T} {\Phi}_{y}$ be kernel matrices, i.e., $({K}_{x})_{i j}={\kappa}\left(\mathbf{x}_{i}, \mathbf{x}_{j}\right)$, where $\mathbf{\kappa}$ is a kernel function, such as a radial basis function (RBF). Then,  (\ref{eq:KCCA2}) can be further simplified to:
\begin{align}
\max _{\mathbf{a}, \mathbf{b}}\ & \mathbf{a}^{T} {K}_{x} {K}_{y} \mathbf{b}\\
s.t.\ &\mathbf{a}^{T} {K}_{x} {K}_{x} \mathbf{a}=1, \mathbf{b}^{T} {K}_{y} {K}_{y} \mathbf{b}=1.\nonumber
\end{align}

We can further add regularizations to the kernel matrices to make them numerically more stable \cite{Hardoon2004}, e.g., replace ${K}_{x}{K}_{x}$ with ${K}_{x}{K}_{x}+r_{x} {K}_{x}$, and ${K}_{y}{K}_{y}$ with ${K}_{y}{K}_{y}+r_{y} {K}_{y}$. The linear coefficient vectors $\mathbf{a}$ and $\mathbf{b}$ can then be solved by the following generalized eigen decomposition problem:
\begin{align}
&\left[\begin{array}{cc}
      \mathbf{0}   & K_{x}K_{y} \\
      K_{y}K_{x} &\mathbf{0}
      \end{array}\right]\left[\begin{array}{c}
                                \mathbf{a} \\
                                \mathbf{b}
                              \end{array}\right]\nonumber\\
=&      \lambda\left[\begin{array}{cc}
                     K_xK_x+r_xK_x & \mathbf{0} \\
                     \mathbf{0} & K_yK_y+r_yK_y
                   \end{array}\right]
                   \left[\begin{array}{c}
                                \mathbf{a} \\
                                \mathbf{b}
                              \end{array}\right]
\end{align}

\subsection{Randomized Nonlinear CCA (RCCA)}

KCCA has very high computational complexity. Lopez-Paz \emph{et al.} \cite{Lopez-Paz2014} handled the nonlinear characteristics of data by performing random nonlinear projection, which greatly reduced the computational difficulty, with little scarification of performance:
\begin{align}
\text{RCCA}\left(X,Y\right) :=\text{CCA}(f_x(X),f_y(Y))\approx \text{KCCA}(X,Y),
\end{align}
where $f_{x}(\cdot)$ represents a mapping function, whose parameters $Q_{x} = \left(\boldsymbol{q}_{x,1}, \boldsymbol{q}_{x,2}, \ldots, \boldsymbol{q}_{x,D}\right) \in \mathbb{R}^{d_{x} \times D}$ are randomly sampled from a given data-independent distribution $p(\boldsymbol{q})$, e.g., a Gaussian distribution. The $D$-dimensional nonlinear random features $Z_{x}$ can be obtained by:
\begin{align}
\mathbf{z}_{x,d} :=&\left[\cos \left(\boldsymbol{q}_{x,d}^{T} \mathbf{x}_{1}+b_{d}\right), \ldots, \cos \left(\boldsymbol{q}_{x,d}^{T} \mathbf{x}_{N}+b_{d}\right)\right]^{T} \in \mathbb{R}^{N\times 1}, \label{eq:zxd} \\
Z_{x}=&\left[\mathbf{z}_{x,1}, \ldots, \mathbf{z}_{x,D} \right]^{T}\in \mathbb{R}^{D\times N},\nonumber
\end{align}
where $\mathbf{b}=[b_{1}, b_{2}, \ldots, b_{D}]^T\in \mathbb{R}^{D\times 1}$ is a randomly generated bias term. The cosine mapping function in (\ref{eq:zxd}) can also be replaced by other nonlinear functions, e.g., sine. As in CCA, the computational complexity of RCCA also increases linearly with the sample size.

\subsection{Locality Preserving CCA (LPCCA)}

In addition to the computational complexity, the global kernelization strategy of KCCA ignores the local linear structure of complex data. Locality preserving CCA (LPCCA) \cite{Sun2007} is another nonlinear CCA extension, which preserves the local linear structure of the data while performing global nonlinear dimensionality reduction.

LPCCA assumes that the corresponding instances of different views should be as close as possible in the common latent space, so it can be expressed in the following equivalent form \cite{Hoegaerts2005}:
\begin{align}
\min _{\mathbf{w}_{x}, \mathbf{w}_{y}}& \sum_{i=1}^{N}\left\|\mathbf{w}_{x}^{T}\left(\mathbf{x}_{i}-\overline{\mathbf{x}}\right)
-\mathbf{w}_{y}^{T}\left(\mathbf{y}_{i}-\overline{\mathbf{y}}\right)\right\|_{2}^{2},\label{eq:wx0} \\
s.t.& \sum_{i=1}^{N}\left\|\mathbf{w}_{x}^{T}\left(\mathbf{x}_{i}-\overline{\mathbf{x}}\right)\right\|^{2}=1, \sum_{i=1}^{n}\left\|\mathbf{w}_{y}^{T}\left(\mathbf{y}_{i}-\overline{\mathbf{y}}\right)\right\|^{2}=1 \nonumber
\end{align}
where $\overline{\mathbf{x}}=\frac{1}{N}\sum_{i=1}^{N}\mathbf{x}_{i}$ and $\overline{\mathbf{y}}=\frac{1}{N}\sum_{i=1}^{N}\mathbf{y}_i$ represent the mean vectors of ${X}$ and ${Y}$, respectively. After some algebraic manipulations, (\ref{eq:wx0}) can be re-expressed as:
\begin{align}
\max_{\mathbf{w}_{x}, \mathbf{w}_{y}} &\mathbf{w}_{x}^{T} \cdot \sum_{i=1}^{N} \sum_{j=1}^{N}\left(\mathbf{x}_{i}-\mathbf{x}_{j}\right)\left(\mathbf{y}_{i}-\mathbf{y}_{j}\right)^{T} \cdot \mathbf{w}_{y} \label{equ:CCA2}\\
s.t. \ &\mathbf{w}_{x}^{T} \cdot \sum_{i=1}^{N} \sum_{j=1}^{N}\left(\mathbf{x}_{i}-\mathbf{x}_{j}\right)\left(\mathbf{x}_{i}-\mathbf{x}_{j}\right)^{T} \cdot \mathbf{w}_{x}=1, \nonumber \\
&\mathbf{w}_{y}^{T} \cdot \sum_{i=1}^{N} \sum_{j=1}^{N}\left(\mathbf{y}_{i}-\mathbf{y}_{j}\right)\left(\mathbf{y}_{i}-\mathbf{y}_{j}\right)^{T} \cdot \mathbf{w}_{y}=1. \nonumber
\end{align}

Let $\text{Nei} \left(\mathbf{x}_i\right)$ be the neighbor set of $\mathbf{x}_{i}$. Define a similarity matrix ${S}_{x}$, whose $ij$-th element is:
\begin{align}
S_{x,i j}=\left\{\begin{array}{ll}
\exp \left(-\left\|\mathbf{x}_{i}-\mathbf{x}_{j}\right\|_{2}^{2} / t_{x}\right),& \mathbf{x}_{j} \in \text{Nei}\left(\mathbf{x}_{i}\right) \\
0, &\text { otherwise }\end{array}\right. ,
\end{align}
where $t_{x}=\sum_{i=1}^{N}\sum_{j=1}^{N} \frac{ 2\left\|\mathbf{x}_{i}-\mathbf{x}_{j}\right\|_{2}^{2}}{N(N-1)}$ represents the mean squared distance between all instances. The similarity matrix ${S}_{y}$ of ${Y}$ can be computed in a similar manner.

Substituting $S_x$ and $S_y$ into (\ref{equ:CCA2}), the global correlation is decomposed into many local linear correlations between the neighboring instances. The objective of LPCCA is:
\begin{align}
\max _{\mathbf{w}_{x}, \mathbf{w}_{y}} &\mathbf{w}_{x}^{T} \cdot \sum_{i=1}^{N} \sum_{j=1}^{N}S_{x,i j}\left(\mathbf{x}_{i}-\mathbf{x}_{j}\right)S_{y,ij}\left(\mathbf{y}_{i}-\mathbf{y}_{j}\right)^{T} \cdot \mathbf{w}_{y}\\  \nonumber
s.t. \ &\mathbf{w}_{x}^{T} \cdot \sum_{i=1}^{N} \sum_{j=1}^{N}S_{x,i j}\left(\mathbf{x}_{i}-\mathbf{x}_{j}\right)S_{x,ij}\left(\mathbf{x}_{i}-\mathbf{x}_{j}\right)^{T} \cdot \mathbf{w}_{x}=1, \\ \nonumber
&\mathbf{w}_{y}^{T} \cdot \sum_{i=1}^{N} \sum_{j=1}^{N}S_{y,i j}\left(\mathbf{y}_{i}-\mathbf{y}_{j}\right)S_{y,ij}\left(\mathbf{y}_{i}-\mathbf{y}_{j}\right)^{T} \cdot \mathbf{w}_{y}=1,
\end{align}
which can be transformed into a generalized eigen decomposition problem:
\begin{align}
&\left[
\begin{array}{cc}
  \mathbf{0} &XS_{xy}Y^T \\
  YS_{yx}X^T & \mathbf{0}
\end{array}\right]\left[\begin{array}{c}
                          \mathbf{w}_x \\
                          \mathbf{w}_y
                        \end{array}\right] \nonumber \\
=&\lambda\left[\begin{array}{cc}
                 XS_{xx}X^T & \mathbf{0} \\
                 \mathbf{0} & YS_{yy}Y^T
               \end{array}\right]
               \left[\begin{array}{c}
                          \mathbf{w}_x \\
                          \mathbf{w}_y
                        \end{array}\right],
\end{align}
where
\begin{align}
{S}_{xy}&={D}_{xy}-{S}_{x}\odot{S}_y,\\
{S}_{yx}&={D}_{yx}-{S}_{y}\odot{S}_x,\\
{S}_{xx}&={D}_{xx}-{S}_{x}\odot{S}_x,\\
{S}_{yy}&={D}_{yy}-{S}_{y}\odot{S}_y,
\end{align}
in which $\odot$ denotes the element-wise product operation, and $D_{xy}$ is a diagonal matrix, with $(D_{xy})_{ii}=\sum_{j=1}^{N}({S}_{x}\odot{S}_y)_{ij}$. $D_{yx}$, $D_{xx}$ and $D_{yy}$ are defined similarly.

\subsection{Discriminative CCA (DisCCA)}

Traditional CCA is unsupervised. In supervised classification, we have label information, which should be taken into consideration to help extract more discriminative features.

Discriminative CCA (DisCCA) is one such approach. It maximizes the within-class similarity and minimizes the between-class similarity. Rearrange $X$ and $Y$ according to the classes:
\begin{align}
&\hat{{X}}=\left[\mathbf{x}_{1}^{(1)}, \ldots, \mathbf{x}_{n_{1}}^{(1)}, \ldots \ldots, \mathbf{x}_{1}^{(c)}, \dots, \mathbf{x}_{n_{c}}^{(c)}\right],\label{equ:Xhat}\\
&\hat{{Y}}=\left[\mathbf{y}_{1}^{(1)}, \ldots, \mathbf{y}_{n_{1}}^{(1)}, \ldots \ldots, \mathbf{y}_{1}^{(c)}, \ldots, \mathbf{y}_{n_{c}}^{(c)}\right],
\end{align}
where $\mathbf{x}_{j}^{(i)}$ and $\mathbf{y}_{j}^{(i)}$ are the $j$-th instance of Class $i$ from the two views.  The objective function of DisCCA is:
\begin{align}
\max _{\mathbf{w}_{x}, \mathbf{w}_{y}}\ &\mathbf{w}_{x}^{T} {C}_{\mathrm{w}} \mathbf{\mathrm{w}}_{y}-\eta \cdot \mathbf{w}_{x}^{T} {C}_{\mathrm{b}} \mathbf{w}_{y} \label{eq:wx}\\
s.t.\ & \mathbf{w}_{x}^{T} \hat{X} \hat{X}^{T} \mathbf{w}_{x}=1, \mathbf{w}_{y}^{T} \hat{Y} \hat{Y}^{T} \mathbf{w}_{y}=1,\nonumber
\end{align}
where $\eta$ is a trade-off parameter, and the within-class similarity matrix ${C}_{\mathrm{w}}$ and between-class similarity matrix ${C}_{\mathrm{b}}$ are defined as:
\begin{align}
&{C}_{\mathrm{w}}=\sum_{i=1}^{c} \sum_{k=1}^{n_{i}} \sum_{l=1}^{n_{j}} \mathbf{x}_{k}^{(i)} \left(\mathbf{y}_{l}^{(j)}\right)^T=\hat{X} {A} \hat{Y}^{T},\\
&{C}_{\mathrm{b}}=\sum_{i=1}^{c} \sum_{j=1 \atop j \neq i}^{c} \sum_{k=1}^{n_{i}} \sum_{l=1}^{n_{j}} \mathbf{x}_{k}^{(i)} \left(\mathbf{y}_{l}^{(j)}\right)^ T=-\hat{X} {A} \hat{Y}^{T},
\end{align}
where $A$ is a block-diagonal matrix:
\begin{align}\label{equ:A}
{A} = \left[\begin{array}{ccccc}
             \mathbf{1}_{n_{1} \times n_{1}} & \quad & \quad & \quad & \quad \\
             \quad & \ddots & \quad & \quad & \quad \\
             \quad & \quad & \mathbf{1}_{n_{i} \times n_{i}} & \quad & \quad \\
             \quad & \quad & \quad & \ddots & \quad \\
             \quad & \quad & \quad & \quad & \mathbf{1}_{n_{c} \times n_{c}}
           \end{array}\right].
\end{align}

Since ${C}_{\mathrm{w}}=-{C}_{\mathrm{b}}$, the optimization is independent of $\eta$, and hence (\ref{eq:wx}) can be rewritten as:
\begin{align}
\max _{\mathbf{w}_{x}, \mathbf{w}_{y}}\ &\mathbf{w}_{x}^{T} \hat{X} {A} \hat{Y}^{T} \mathbf{w}_{y}\\   \nonumber
s.t.\ &\mathbf{w}_{x}^{T} \hat{X} \hat{X}^{T} \mathbf{w}_{x}=1, \mathbf{w}_{y}^{T} \hat{Y} \hat{Y}^{T} \mathbf{w}_{y}=1,
\end{align}
whose solution is similar to (\ref{equ:CCA}) and can also be expressed as a generalized eigen decomposition problem:
\begin{align}
\begin{bmatrix}{\mathbf{0}}&{\tilde{\Sigma}_{xy}} \\ {\tilde{\Sigma}_{yx}}&{\mathbf{0}}\end{bmatrix}
\begin{bmatrix}{\mathbf{w}_{\mathrm{x}}} \\ {\mathbf{w}_{\mathrm{y}}}\end{bmatrix}=\lambda
\begin{bmatrix}{\hat{\Sigma}_{xx}} & {\mathbf{0}} \\ {\mathbf{0}} & {\hat{\Sigma}_{yy}}\end{bmatrix}
\begin{bmatrix}{\mathbf{w}_{x}} \\ {\mathbf{w}_{y}}\end{bmatrix},
\end{align}
where $\tilde{\Sigma}_{xy}=\frac{1}{N}\hat{X}{A} \hat{Y}^{T}$ and $\tilde{\Sigma}_{yx}=\frac{1}{N}\hat{Y} {A} \hat{X}^{T}$.

\subsection{Multi-view Linear Discriminant Analysis (MLDA)}

LDA \cite{Fisher1936} is a supervised algorithm for a single view that minimizes the within-class variance and maximizes the between-class variance. Multi-view linear discriminant analysis (MLDA) \cite{Sun2016} combines LDA and CCA, which not only ensures the discriminative ability within a single view, but also maximizes the correlation between different views.

Define a between-class scatter matrix $S_{b}=\frac{1}{N} \hat{{X}} {W} \hat{X}^{T}$, where $\hat{X}$ is given in (\ref{equ:Xhat}), $W = diag({W}_{1},{W}_2,\dots, {W}_{c})$, with all elements in ${W}_{i}\in \mathbb{R}^{n_{i}\times n_{i}}$ equal $\frac{1}{n_i}$. Then, for a single view, the objective function of LDA is:
\begin{align}
\max _{\mathbf{w}}\ & \mathbf{w}^{T} {S}_{\mathrm{b},x} \mathbf{w}\\ \nonumber
s.t.\ & \mathbf{w}^{T}\hat{\Sigma}_{xx} \mathbf{w}=1.
\end{align}

Integrating with (\ref{equ:CCA}), the objective function of MLDA is:
\begin{align}\label{equ:MLDA}
\max _{\mathbf{w}_{x}, \mathbf{w}_{y}}\ & \mathbf{w}_{x}^{T} {S}_{\mathrm{b},x} \mathbf{w}_{x}+\mathbf{w}_{y}^{T} {S}_{\mathrm{b},y} \mathbf{w}_{y}+\eta \mathbf{w}_{x}^{T} \Sigma_{xy} \mathbf{w}_{y} \\  \nonumber
s.t.\ & \mathbf{w}_{x}^{T} \hat{\Sigma}_{xx} \mathbf{w}_{x}+\sigma\mathbf{w}_{y}^{T} \hat{\Sigma}_{yy} \mathbf{w}_{y}=1,
\end{align}
where $\eta$ is a trade-off parameter, and $\sigma=\frac{tr(\hat{\Sigma}_{xx})}{tr(\hat{\Sigma}_{yy})}$. The constraint in (\ref{equ:MLDA}) is in the form of summation to ensure a closed-form solution can be obtained. Using the Lagrangian multiplier, (\ref{equ:MLDA}) can also be solved by generalized eigen decomposition:
\begin{align}
\begin{bmatrix}{S_{\mathrm{b},x}} & {\eta {\Sigma}_{x y}} \\ {\eta {\Sigma}_{y x}} & {S_{\mathrm{b},y}}\end{bmatrix}
\begin{bmatrix}{\mathbf{w}_{\mathrm{x}}} \\ {\mathbf{w}_{\mathrm{y}}}\end{bmatrix}=\lambda
\begin{bmatrix}{\hat{\Sigma}_{xx}} & {\mathbf{0}} \\ {\mathbf{0}} & {\hat{\Sigma}_{yy}}\end{bmatrix}
\begin{bmatrix}{\mathbf{w}_{\mathbf{x}}} \\ {\mathbf{w}_{\mathbf{y}}}\end{bmatrix}.
\end{align}

\subsection{Multi-view Uncorrelated Linear Discriminant Analysis (MULDA)}

Canonical vectors $\{\left(\mathbf{w}_{x,k}, \mathbf{w}_{y,k}\right)\}_{k=1}^K$, calculated by MLDA, are correlated, which means the projected canonical variables contain redundant information. Multi-view uncorrelated linear discriminant analysis (MULDA) \cite{Sun2016} aims to eliminate the information redundancy of the canonical variables using uncorrelated linear discriminant analysis (ULDA) \cite{Jin2001}.

ULDA imposes orthogonal constraints to LDA, and is solved recursively. Each time a projection vector $\mathbf{w}_k$ is calculated to be orthogonal to the $k-1$ projection vectors that have been obtained:
\begin{align}
\max _{\mathbf{w}} \ & \mathbf{w}^{T} {S}_{\mathrm{b},x} \mathbf{w}\\   \nonumber
s.t.\ & \mathbf{w}_{k}^{T} \hat{\Sigma}_{xx} \mathbf{w}_{i}=0,\quad i=1,2, \ldots, k-1.
\end{align}

Integrating ULDA with MLDA, the objective function of MULDA can be written as:
\begin{align}
\max _{\mathbf{w}_{x,k}, \mathbf{w}_{y,k}}\ & \mathbf{w}_{x,k}^{T} {S}_{\mathrm{b},x} \mathbf{w}_{x,k}+\mathbf{w}_{y,k}^{{T}} {S}_{\mathrm{b},y} \mathbf{w}_{y,k}+\eta \mathbf{w}_{x,k}^{T} \Sigma_{x y} \mathbf{w}_{y,k} \\ \nonumber
s.t.\ & \mathbf{w}_{x,k}^{T} \hat{\Sigma}_{xx} \mathbf{w}_{x,k}+\sigma\mathbf{w}_{y,k}^{T} \hat{\Sigma}_{yy} \mathbf{w}_{y,k}=1, \\ \nonumber
&\mathbf{w}_{x,k}^{T} \hat{\Sigma}_{xx} \mathbf{w}_{x,i}=0, \mathbf{w}_{y,k}^{T} \hat{\Sigma}_{yy} \mathbf{w}_{y,i}=0, \\  \nonumber
& i=1,\ldots,k-1.
\end{align}
The $k$-th pair of canonical vectors, $\left(\mathbf{w}_{x,k},\mathbf{w}_{y,k}\right)$, is the leading eigenvector of
\begin{align}
\begin{bmatrix} {P_{x}}& {\mathbf{0}} \\ {\mathbf{0}} & {P_{y}} \end{bmatrix}
\begin{bmatrix} {S_{\mathrm{b},x}} & {\eta {\Sigma}_{x y}} \\ {\eta {\Sigma}_{y x}} & {S_{\mathrm{b},y}} \end{bmatrix}
\begin{bmatrix} {\mathbf{w}_{x,k}} \\ {\mathbf{w}_{y,k}}\end{bmatrix}=\lambda
\begin{bmatrix} {\hat{\Sigma}_{xx}} & {\mathbf{0}} \\ {\mathbf{0}} & {\hat{\Sigma}_{yy}}\end{bmatrix}
\begin{bmatrix} {\mathbf{w}_{x,k}} \\ {\mathbf{w}_{y,k}}\end{bmatrix},
\end{align}
where
\begin{align}
&{P}_{x}={I}-\hat{\Sigma}_{xx} {D}_{x}^{T}\left({D}_{x} \hat{\Sigma}_{xx} {D}_{x}^{T}\right)^{-1} {D}_{x}, \\
&{D}_{x}=\left[\mathbf{w}_{x,1}, \mathbf{w}_{x,2}, \cdots, \mathbf{w}_{x,k-1}\right]^{T}.
\end{align}
${P}_{y}$ and ${D}_{y}$ are defined similarly.

Unlike DisCCA, MLDA and MULDA only focus on the within-view class scatters, without considering the between-view class scatter. Sun \emph{et al.} \cite{Sun2016} proposed that CCA in MLDA and MULDA can be replaced by DisCCA, to also consider the between-view class scatter. Moreover, MLDA and MULDA can also be extended using kernels to consider the nonlinear relationship between different views.

\subsection{Deep CCA (DCCA) and Discriminative DCCA (DisDCCA)}

\begin{figure}[htbp] \centering
\includegraphics[height=.7\linewidth,clip]{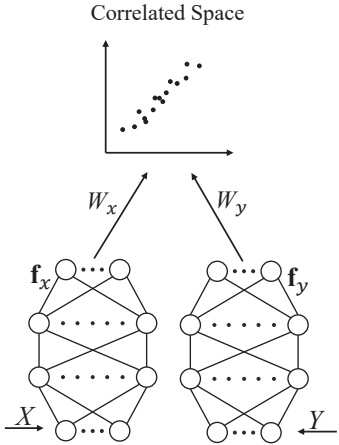}
\caption{Schematic diagram of DCCA. Two deep neural networks $\mathbf{f}_{x}$ and $\mathbf{f}_{y}$ first extract nonlinear features from $X$ and $Y$, respectively, and then a linear CCA is applied for correlation analysis. } \label{fig:DCCA}
\end{figure}

Deep canonical correlation analysis (DCCA) \cite{Andrew2013}, shown in Fig.~\ref{fig:DCCA}, first extracts nonlinear features through deep neural networks (DNNs), and then uses linear CCA to calculate the canonical matrices.

Let $\mathbf{f}_x$ and $\mathbf{f}_y$ be two DNNs, and ${H}_{x}=\mathbf{f}_{x}(X)$ and ${H}_{y}=\mathbf{f}_{y}(Y)$ be their outputs, respectively. We already know that the total correlation of the $K$ canonical variables equals the sum of the first $K$ singular values of matrix ${T}=\hat{\Sigma}_{x x}^{-1/2}{\Sigma}_{x y}\hat{\Sigma}_{y y}^{-1/2}$ from \ref{subsection:CCA}. Define
\begin{align} \label{equ:T}
\hat{T}=&\left(\frac{1}{N}H_{x}H_{x}^{T}+r_{x}I\right)^{-\frac{1}{2}}\nonumber\\
& \left(\frac{1}{N}H_{x}H_{y}^{T}\right)\left(\frac{1}{N}H_{y}H_{y}^{T}+r_{y}I\right)^{-\frac{1}{2}}.
\end{align}
The objective function of DCCA is:
\begin{align}
\max_{\mathbf{f}_{x}, \mathbf{f}_{y}, \mathbf{w}_{x}, \mathbf{w_{y}}}&\sum_{k=1}^{K} \sigma_{k}(\hat{T})\label{equ:DCCA} \\
s.t.\ &\mathbf{w}_{x}{\left(\frac{1}{N}H_{x}H_{x}^{T}+r_{x}I\right)}\mathbf{w}_x^{T}=1,\nonumber \\
& \mathbf{w}_{y}{\left(\frac{1}{N}H_{y}H_{y}^{T}+r_{y}I\right)}\mathbf{w}_y^{T}=1,\nonumber
\end{align}
where $\sigma_{k}(\hat{T})$ denotes the $k$-th largest singular value of $\hat{T}$.

Andrew \emph{et al.} \cite{Andrew2013} employed a full-batch optimization algorithm (L-BFGS) to optimize (\ref{equ:DCCA}), because the covariance matrix of the entire training set needs to be computed. If the dataset is divided into small batches, then the covariance matrix computed from each batch may not be accurate. However, full-batch optimization is both memory-hungry and time-consuming, especially when the dataset is large. Wang \emph{et al.} \cite{Wang2015Unsupervised} proposed that with a large batch size, the instances in each batch can be sufficient for estimating the covariance matrix, and the efficient mini-batch stochastic gradient descent approach can be used.

In experiments, the network outputs on the training set were first obtained, and then a linear CCA was used to obtain the canonical matrices ${W}_{x}$ and ${W}_y$. On the test set, ${W}_{x}^{T}\mathbf{f}_{x}(\cdot)$ and ${W}_{y}^{T}\mathbf{f}_{y}(\cdot)$ were the final inputs to a classifier.

Similar to DCCA, discriminative deep canonical correlation analsis (DisDCCA) \cite{Elmadany2016} is a DNN-based extension of discriminative CCA. There are two major differences in their implementations. First, when calculating the loss of each batch, the instances are first rearranged according to their classes, and then $\frac{1}{N}H_{x}H_y^T$ in (\ref{equ:T}) is replaced by $\frac{1}{N} {H}_{x}{A} {H}_{y}^{T}$, where $A$ is defined in (\ref{equ:A}). Second, after the model is trained, DisCCA is used to obtain the canonical matrices ${W}_{x}$ and ${W}_{y}$.

\subsection{Deep Canonically Correlated Autoencoders (DCCAE) and Discriminative DCCAE (DisDCCAE)}

\begin{figure}[htbp] \centering
\includegraphics[height=.7\linewidth,clip]{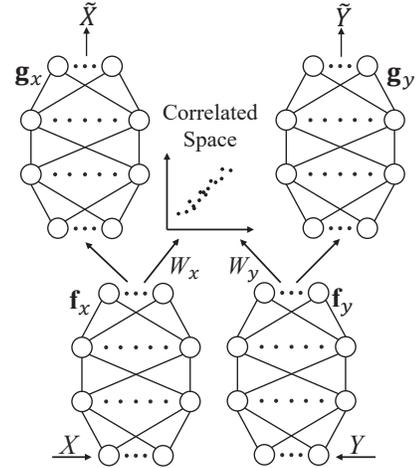}
\caption{Schematic diagram of DCCAE. Two deep neural networks $\mathbf{f}_{x}$ and $\mathbf{f}_{y}$ first extract nonlinear features from $X$ and $Y$, respectively, and then a linear CCA is applied for correlation analysis. $\mathbf{g}_{x}$ and $\mathbf{g}_{y}$ are two decoders for reconstructing View $X$ and View $Y$, respectively. }\label{fig:DCCAE}
\end{figure}

Deep canonically correlated autoencoders (DCCAE) \cite{Wang2015DCCAE} improves DCCA, by using autoencoders to make sure the information captured by $\mathbf{f}_x$ and $\mathbf{f}_y$ can also accurately reconstruct the original $X$ and $Y$. Fig.~\ref{fig:DCCAE} shows the DCCAE model structure, where $\mathbf{g}_{x}$ and $\mathbf{g}_{y}$ represent the decoder networks for reconstructing $X$ and $Y$, respectively. Therefore, two reconstruction errors are incorporated into (\ref{equ:DCCA}):
\begin{align}
\max_{\mathbf{f}_{x}, \mathbf{f}_{y},\mathbf{g}_{x}, \mathbf{g}_{y}, \mathbf{w}_{x}, \mathbf{w_{y}}}\ &\sum_{k=1}^{K} \sigma_{k}\left(\hat{T}\right)
-\frac{\lambda}{N}\left(\left\|{X}-\mathbf{g}_{x}\left(\mathbf{f}_{x}\left({X}\right)\right)\right\|^{2}_{F} \right. \nonumber\\  &
\left.+ \left\|{Y}-\mathbf{g}_{y}\left(\mathbf{f}_{y}\left({Y}\right)\right)\right\|^{2}_{F}\right) \label{equ:DCCAE}\\
s.t.\ &\mathbf{w}_{x}{(\frac{1}{N}H_{x}H_{x}^{T}+r_{x}I)}\mathbf{w}_x^{T}=1,\nonumber \\
& \mathbf{w}_{y}{(\frac{1}{N}H_{y}H_{y}^{T}+r_{y}I)}\mathbf{w}_y^{T}=1.\nonumber
\end{align}

The test procedure is similar to DCCA, except that ${W}_{x}^{T}\mathbf{f}_{x}(\cdot)$ and ${W}_{y}^{T}\mathbf{f}_{y}(\cdot)$ are used as inputs to a classifier. The advantage of DCCAE is that the autoencoders can alleviate the over-fitting of the model.

Since the first term of (\ref{equ:DCCAE}) is identical to that in (\ref{equ:DCCA}), extending DCCAE to discriminative DCCAE (DisDCCAE) is similar to extending DCCA to DisDCCA. However, to our knowledge, no one has done that before. We included both DCCAE and DisDCCAE in our experiments.

\subsection{Variational CCA (VCCA) and VCCA-private}

Bach and Jordan \cite{Bach2005} explained CCA from the perspective of a probabilistic latent variable model. It assumes that the instances ${\mathbf{x}}$ and ${\mathbf{y}}$ from two views are independently conditioned on the multivariate latent variable ${\mathbf{z}}\in \mathbb{R}^{d_{z}\times 1}$, and CCA aims to maximize the joint probability distribution of $\mathbf{x}$ and $\mathbf{y}$:
\begin{align} p(\mathbf{x}, \mathbf{y}, \mathbf{z}) &=p(\mathbf{z}) p(\mathbf{x} | \mathbf{z}) p(\mathbf{y} | \mathbf{z}), \\ p(\mathbf{x}, \mathbf{y}) &=\int p(\mathbf{x}, \mathbf{y}, \mathbf{z}) d \mathbf{z}.
\end{align}

\begin{figure*}[htbp] \centering
\subfigure[VCCA]{\label{fig:DVCCA} \includegraphics[height=.35\linewidth,trim = 120 5 120 10, clip]{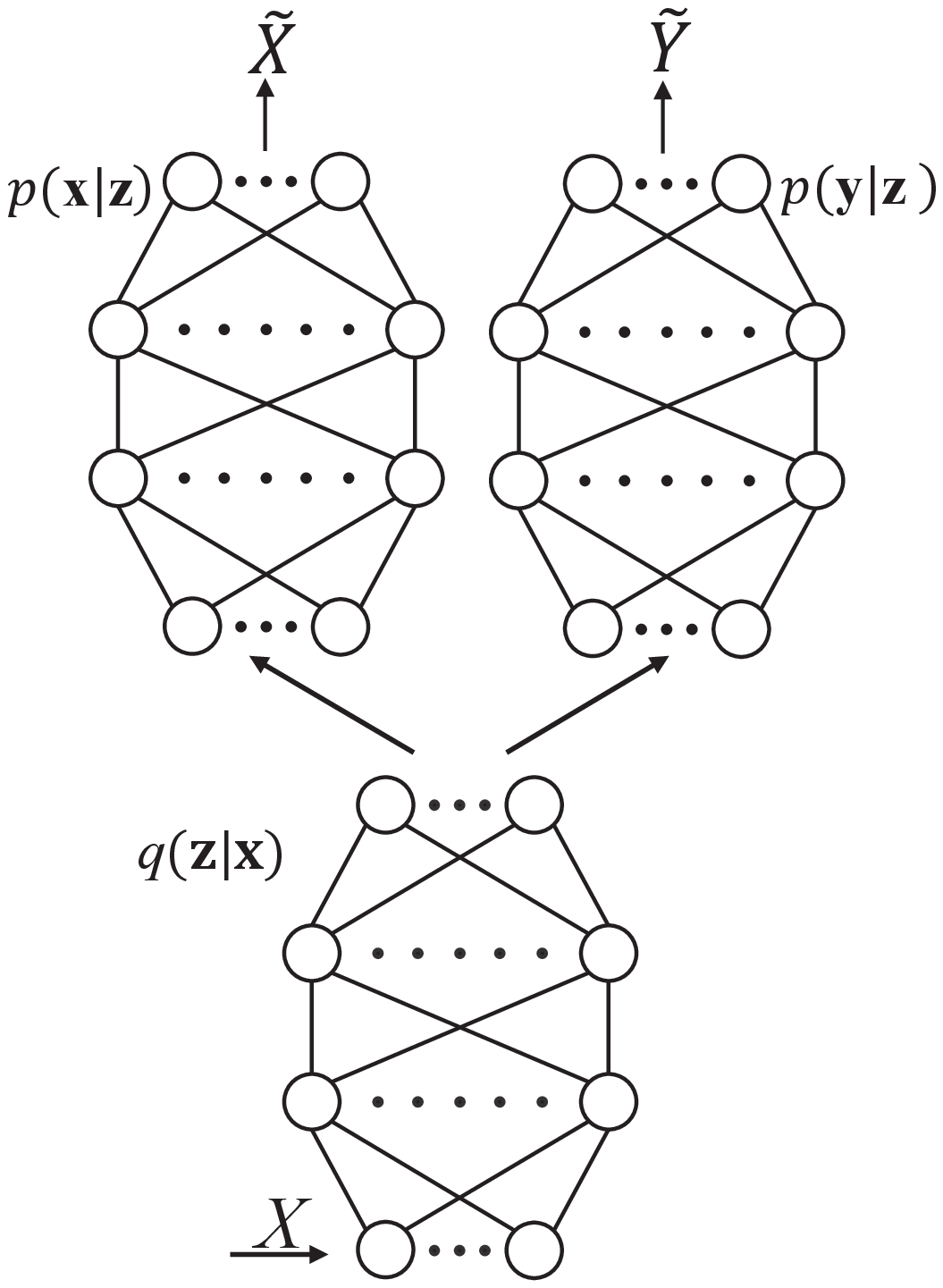}}
\subfigure[VCCA-private]{\label{fig:DVCCAP} \includegraphics[height=.35\linewidth,trim = 100 5 100 10, clip]{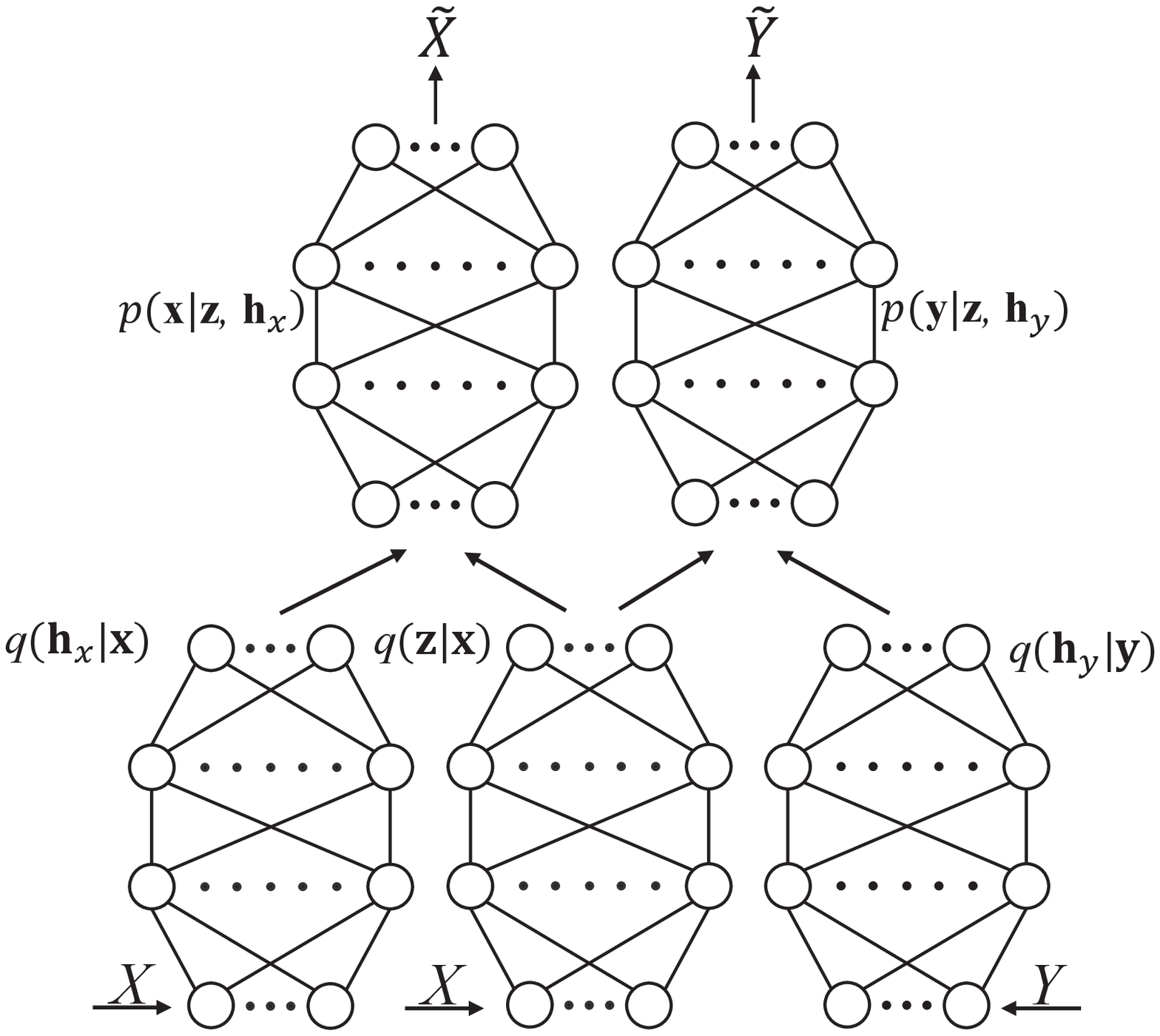}}\centering
\caption{Schematic diagrams of DVCCA and DVCCA-private. The encoders $q(\cdot |x)$ and $q(\cdot |y)$ are used to fit the mean and covariance of the input, and the $p(x|\cdot)$ and $p(y| \cdot)$ are decoders for reconstructing input data. }
\end{figure*}

Deep variational canonical correlation analysis (VCCA) \cite{Wang2016}, shown in Fig.~\ref{fig:DVCCA}, was extended from variational autoencoders (VAE) \cite{Kingma2013}. First, the mean vector $\boldsymbol{\mu}_{i}$ and the diagonal covariance matrix $\boldsymbol{\Sigma}_{i}$ of $\mathbf{x}_i$ are calculated by the encoder $\mathbf{f}_{x}$. Then, $L$ instances, $\{\mathbf{z}_{i}^{(l)}\}_{l=1}^L$, are randomly sampled from the distribution $\mathcal{N}\left(\boldsymbol{\mu}_{i}, \boldsymbol{\Sigma}_{i}\right)$. Finally, decoders $\mathbf{g}_{x}$ and $\mathbf{g}_{y}$ reconstruct instances $\mathbf{x}$ and $\mathbf{y}$, respectively.

The objective function of VCCA is:
\begin{align}
\min _{\mathbf{f}_x,\mathbf{g}_x,\mathbf{g}_y}& \frac{1}{N} \sum_{i=1}^{N} D_{K L}\left(q( \mathbf{z}_i | \mathbf{x}_i)| | p\left(\mathbf{z}_{i}\right) \right)+ \\ \nonumber
&\frac{\lambda}{N L} \sum_{i=1}^{N} \sum_{l=1}^{L} \left(\log p\left(\mathbf{x}_{i} | \mathbf{z}_{i}^{(l)}\right)+\log p\left(\mathbf{y}_{i} | \mathbf{z}_{i}^{(l)}\right)\right),
\end{align}
where the first term denotes the Kullback-Leibler (KL) divergence between the posterior distributions $q(\mathbf{z}_{i}|\mathbf{x}_{i})$ and $ p\left(\mathbf{z}_{i}\right)$, and the latter two terms denote the expectations of the log-likelihood under the approximate posterior distributions, which are equivalent to the reconstruction error. In testing, the mean vector $\boldsymbol{\mu}_{i}$  is used as the input features.

VCCA only takes the common latent variables $\mathbf{z}$ into consideration, which may not be sufficient to describe all information in $X$ and $Y$. In order to more adequately capture the private information of the views, Wang \emph{et al.} \cite{Wang2016} further proposed VCCA-private, shown in Fig.~\ref{fig:DVCCAP}. Its probabilistic model is defined as:
\begin{align}
p(\mathbf{x}, \mathbf{y}, \mathbf{z}, \mathbf{h}_{x}, \mathbf{h}_{y}) &=p(\mathbf{z})p(\mathbf{h}_{x})p(\mathbf{h}_{y}) p(\mathbf{x} | \mathbf{z},\mathbf{h}_{x}) p(\mathbf{y} | \mathbf{z}, \mathbf{h}_{y}), \\
p(\mathbf{x}, \mathbf{y}) &=\iiint p(\mathbf{x}, \mathbf{y}, \mathbf{z},\mathbf{h}_{x}, \mathbf{h}_{y} ) d \mathbf{z} d \mathbf{h}_{x} d \mathbf{h}_{y}.
\end{align}

The corresponding objective function of VCCA-private is:
\begin{align}
\min_{\mathbf{f}_{x,h},\mathbf{f}_{x,z},\mathbf{f}_{y,h}, \mathbf{g}_{x}, \mathbf{g}_{y}} &\frac{1}{N} \sum_{i=1}^{N}
(D_{K L}\left(q\left( \mathbf{z}_{i} | \mathbf{x}_{i}\right)| | p\left(\mathbf{z}_{i}\right) \right) \\
&+D_{K L}\left(q\left( \mathbf{h}_{x,i} | \mathbf{x}_{i}\right)| | p\left(\mathbf{h}_{x,i}\right) \right)\nonumber \\
&+D_{K L}\left(q\left( \mathbf{h}_{y,i} | \mathbf{y}_{i}\right)| | p\left(\mathbf{h}_{y,i}\right) \right)
)\nonumber \\
&+ \frac{\lambda}{N L} \sum_{i=1}^{N} \sum_{l=1}^{L} \left(\log p\left(\mathbf{x}_{i} | \mathbf{z}_{i}^{(l)},\mathbf{h}_{x,i}^{(l)}\right)\right.\nonumber \\
&+\left.\log p\left(\mathbf{y}_{i} | \mathbf{z}_{i}^{(l)},\mathbf{h}_{y,i}^{(l)}\right)\right).\nonumber
\end{align}

In testing, the mean vectors output by the three encoders are stacked as new features. According to \cite{Wang2016}, VCCA-private is more effective than VCCA.

\subsection{Summary}

This section has introduced the traditional CCA and its six nonlinear extensions, three supervised extensions, and one sparse extension. Three ideas were used in the nonlinear extensions, namely, kernel-based KCCA, locality preserving LPCCA, and DNN-based DCCA (DCCAE, DVCCA). The first two are more suitable for small datasets. The supervised extensions (DisCCA, MLDA and MULDA) are closely related to LDA, because they all take the between-class and within-class scatter matrices into consideration.

\section{CCA-Based MVL for More than Two Views}\label{sec:MultiViewCCA}

This section summarizes five representative CCA approaches for more than two views, which are SUMCOR-GCCA, MAXVAR-GCCA, LS-CCA, TCCA, and DGCCA.

\subsection{SUMCOR-GCCA}

CCA maximizes the correlation between two views. For more than two views, a natural extension is to maximize the sum of the pairwise correlations \cite{Horst1961} \cite{Kettenring1971}, which is the idea of SUMCOR-GCCA.

Let $\left\{{X}_{j} \in \mathbb{R}^{d_{j} \times N}\right\}_{j=1}^J$ be a dataset containing $J$ mean-zero views, where $d_{j}$ is the feature dimensionality of View $j$. The objective function of SUMCOR-GCCA is:
\begin{align}
\max _{\left\{\mathbf{w}_{i}\right\}_{i=1}^{J}}\ & \sum_{i=1}^{J}\sum_{j=1}^{J}\mathbf{w}_{i}^{T}{X}_{i}{X}_{j}^{T}\mathbf{w}_{j}\\ \nonumber
s.t.\ &\mathbf{w}_{j}^{T}X_{j}X_{j}^{T}\mathbf{w}_{j}=1,\quad j=1,\cdots,J.
\end{align}

There is no closed-form solution for $\mathbf{w}_{j}$. Therefore, we followed \cite{Asendorf2015} and solved it using the Manopt \cite{Manopt} package\footnote{www.manopt.org}.

\subsection{MAXVAR-GCCA}

GCCA \cite{Carroll1968} assumes that each view can be generated from a set of multivariate latent variables ${G}= [\mathbf{g}_{1}, \mathbf{g}_{2},\ldots, \mathbf{g}_{N}]^{T} \in \mathbb{R}^{N \times K}$, in which all views are correlated. Its objective function is:
\begin{align}
\min_{G, \left\{W_{j}\right\}_{j=1}^{J}} \sum_{j=1}^{J}\left\|G-X_{j}^{T} W_{j}\right\|_{F}^{2} \quad  s.t.\ G^{T}G=I,
\end{align}
where ${W}_{j}\in \mathbb{R}^{d_{j} \times K}$ is the canonical matrix of View $j$, and the constraint guarantees that $G$ is a unit orthogonal matrix.

This objective is similar to a least squares problem. In order to avoid memory problems for large datasets, following \cite{Rastogi2015,Arora2012}, an SVD is first used to obtain a rank-$m$ approximation of each view:
\begin{align}
{X}_{j} \approx {U}_{j}{S}_{j}{V}_{j}^{T},\quad  j=1,\cdots,J, \label{eq:Xj}
\end{align}
where ${S}_{j}\in \mathbb{R}^{m \times m}$ is a diagonal matrix composed of the $m$ largest singular values, and ${U}_{j}\in {R}^{d_{j} \times m}$ and ${V}_{j}\in \mathbb{R}^{N \times m}$ are the corresponding left and right singular matrices, respectively.

$G$ then consists of the $K$ leading eigenvectors of matrix ${M}=\tilde{M}\tilde{M}^{T}$, where
\begin{align}
\tilde{M}&=\left[V_{1}T_{1},\cdots,V_{J}T_{J}\right]\in \mathbb{R}^{N\times mJ}.
\end{align}
It has been shown \cite{Rastogi2015} that the diagonal matrix $T_{j}$ satisfies $T_{j}T_{j}^{T}=S_{j}\left(S_{j}^{T} S_{j}+r_{j} I\right)^{-1} S_{j}^{T}$. Hence, given $S_j$ in (\ref{eq:Xj}), $T_j$ can be easily computed.

Once $G$ is obtained, $W_j$ can be computed as:
\begin{align}
W_{j}&=\left(X_{j} X_{j}^{T}+r_{j}I\right)^{-1} {X}_{j} G.
\end{align}

Benton \emph{et al.} \cite{Benton2016} also proposed view-weighted GCCA to consider the varying importance of different views.

\subsection{Least Squares based Generalized CCA (LS-CCA)}

LS-CCA \cite{Via2007} aims to minimize the distances among the canonical variables, so that different views are maximally overlapping with each other after mapping. Its objective function is:
\begin{align}
\min  _{\left\{\mathbf{w}_{j}\right\}_{j=1}^{J}} \ & \frac{1}{2 J(J-1)}\sum_{i=1}^{J}\sum_{j=1}^{J}\left\|X_{i}^{T} \mathbf{w}_{i}-X_{j}^{T} \mathbf{w}_{j}\right\|_{2}^{2} \\ \nonumber
s.t.\ &\frac{1}{J} \sum_{j=1}^{J} \mathbf{w}_{j}^{T}XX^{T} \mathbf{w}_{j}=1.
\end{align}

The $k$-th stacked canonical vector $\mathbf{w}^{\left(k\right)}=[\mathbf{w}_{1}^{\left(k\right)};\cdots; \mathbf{w}_{J}^{\left(k\right)}]$ can be obtained by performing a generalized eigen decomposition:
\begin{align}
\frac{1}{J-1}(R-D) \mathbf{w}^{(k)}=\lambda^{(k)} D \mathbf{w}^{(k)},
\end{align}
where
\begin{align}
R=
\begin{bmatrix}
\Sigma_{11} & \cdots & \Sigma_{1J} \\
\vdots & \ddots & \vdots \\
\Sigma_{J1} & \cdots & \Sigma_{JJ}
\end{bmatrix},\quad
D=
\begin{bmatrix}
\hat{\Sigma}_{11} & \cdots & \mathbf{0} \\
\vdots & \ddots & \vdots \\
\mathbf{0} & \cdots & \hat{\Sigma}_{JJ}
\end{bmatrix}.
\end{align}

Since the computational cost of this solution is high, LS-CCA can also be solved in an iterative manner, e.g., using partial least squares (PLS) \cite{Via2007}, so that the canonical vectors are obtained directly without using low rank approximation. Via \emph{et al.} \cite{Via2007} showed that LS-CCA and MAXVAR-GCCA are equivalent. Moreover, when $J=2$, LS-CCA degrades to CCA.

\subsection{Tensor CCA (TCCA)}

Traditional CCA optimizes the pairwise view correlation only, and ignores high-order statistics. TCCA \cite{Luo2015} directly maximizes the correlation of all views, using a high-order covariance tensor. Its objective function is:
\begin{align}
\max _{\left\{\mathbf{w}_{j}\right\}_{j=1}^{J}}\ & \rho\left(\mathbf{w}_{1}^{T}X_{1}, \mathbf{w}_{2}^{T}X_{2}, \ldots, \mathbf{w}_{J}^{T}X_{J}\right)\\ \nonumber
s.t.\  & \mathbf{w}_{j}^{T}XX^{T}\mathbf{w}_{j}=1, \ j=1, \ldots, J,
\end{align}
where $\rho$ denotes the correlation among the $J$ views. Luo \emph{et al.} \cite{Luo2015} showed that
\begin{align}\label{equ:TCCA}
\rho\left(\mathbf{w}_{1}^{T}X_{1}, \ldots, \mathbf{w}_{J}^{T}X_{J}\right)=\mathcal{C}_{1,2, \ldots ,J} \times_{1} \mathbf{w}_{1}^{T} \ldots \times_{J} \mathbf{w}_{J}^{T},
\end{align}
where $\times_{j}$ is the $j$-mode product, and $\mathcal{C}_{1,2, \ldots, J}$ denotes a $d_{1} \times d_{2} \times \cdots \times d_{J}$ covariance tensor. Let $\circ$ be the tensor product. Then,
\begin{align}
\mathcal{C}_{1,2, \ldots, J}=\frac{1}{N} \sum_{i=1}^{N} \mathbf{x}_{1 i} \circ \mathbf{x}_{2 i} \circ \ldots \circ \mathbf{x}_{J i},
\end{align}
where $\mathbf{x}_{ji}$ ($j=1,...,J$) denotes the $i$-th instance of View $j$. Let $\mathbf{u}_{j}=\hat{\Sigma}_{j j}^{1 / 2} \mathbf{w}_{j}$, and
\begin{align}
\mathcal{M}=\mathcal{C}_{1,2, \ldots, J} \times_{1} \hat{\Sigma}_{11}^{-1 / 2} \times_{2} \hat{\Sigma}_{22}^{-1 / 2} \ldots \times_{J} \hat{\Sigma}_{J J}^{-1 / 2}.
\end{align}
Then, (\ref{equ:TCCA}) can be rewritten as:
\begin{align}
\max_ {\left\{\mathbf{u}_{j}\right\}_{j=1}^{J}}\ & \mathcal{M} \times_{1} \mathbf{u}_{1}^{T} \times_{2} \mathbf{u}_{2}^{T} \ldots \times_{J} \mathbf{u}_{J}^{T} \\ \nonumber
s.t.\ & \mathbf{u}_{j}^{T} \mathbf{u}_{j}=1, \quad j=1, \ldots, J.
\end{align}

This problem can be solved by decomposing $\mathcal{M}$ into $K$ best rank-1 approximations $\left[\mathbf{u}_{1}^{k}, \mathbf{u}_{2}^{k}, \cdots, \mathbf{u}_{J}^{k}\right]_{k=1}^{K}$ with alternating least squares, and then stacking the corresponding vectors to obtain the canonical vectors $W_{j}=\left[\hat{\Sigma}_{jj}^{-1/2}\mathbf{u}_{j}^{1}, \cdots, \hat{\Sigma}_{jj}^{-1/2}\mathbf{u}_{j}^{K}\right]$ of View $j$.

\subsection{Deep Generalized CCA (DGCCA)}

\begin{figure}[htpb] \centering
\includegraphics[width=.95\linewidth,clip]{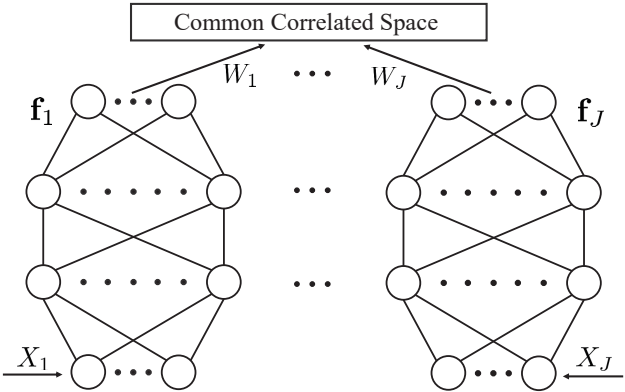}
\caption{A Schematic diagram of DGCCA for $J$ views. Each view $X_{j}, j=1,\cdots,J$, first passes through a multi-layer neural network $\mathbf{f}_{j}, j=1,\cdots,J$, to obtain nonlinear features, and then GCCA is applied to project them onto a common correlated space.}
\label{fig:DGCCA}
\end{figure}

DGCCA \cite{Benton2017}, shown in Fig.~\ref{fig:DGCCA}, is a DNN-based nonlinear extension of GCCA. Each view first passes through a multi-layer perceptron neural network to obtain nonlinear features $H_{j}=\mathbf{f}_{j}\left(X_{j}\right)$, $j=1,\cdots,J$. The objective of DGCCA is defined as the sum of the reconstruction errors between the canonical variables $X_{j}^{T}W_{j} $ and the common representation $G$:
\begin{align}
\min_{\left\{\mathbf{f}_{j}\right\}_{j=1}^{J}} \ \sum_{j=1}^{J}\left\|G-\mathbf{f}_{j}\left(X_{j}\right)^{T} {W}_{j}\right\|_{F}^{2} \quad  s.t.\ G^{T}G=I.
\end{align}

In training, the network's outputs for a mini-batch of instances are used as the inputs to GCCA, and then the reconstruction error between each view and the common representation is calculated to update the corresponding network parameters. In testing, the common representation $G$ or the stacked canonical variables $\left[{X}_{1}^{T}{W}_{1},\cdots,X_{J}^{T}W_{J}\right]$ can be used as the input to a classifier.

DGCCA is unsupervised. To utilize the label information, Benton \emph{et al.} \cite{Benton2017} added the labels' one-hot coding matrix as an additional view.

\subsection{Summary}

This section has introduced five representative CCA approaches for more than two views. To our knowledge, TCCA is currently the only approach to directly calculate the correlation among more than two views. All other approaches perform the computation indirectly.

\section{Application}\label{sec:application}

As a well-known approach for analysing correlations between two sets of variables, CCA and its extensions have been used in a variety of machine learning tasks. According to the type of problems, they can be divided into three main categories: multi-view recognition tasks, cross-modal retrieval/classificaion, and multi-view embedding.

\subsection{Multi-view Recognition Tasks}

In these CCA-based MVL applications, recognition tasks generally refers to supervised or unsupervised classification or regression. These methods have a wide range of applications in computer vision. To name a few, face recognition were achieved by treating the Daubechies wavelet transformed low-frequent images and the original images as two different views \cite{Sun2008, Sun2016}. Zheng \emph{et al.} \cite{Zheng2006} recognized expression of facial images by using the psychologist's semantic rating as the second view. To improve the performance of human gender recognition, Shan \emph{et al.} \cite{Shan2008} fused gait and face modalities at feature level with CCA.  Another challenging problem in computer version is pose estimation, which is try to compute the pose parameters of an image in the three-dimensional scene. A representative approach is first to construct connections between training images with pose parameters by using CCA-based MVL approaches \cite{Sun2007}, and then the pose parameters were estimated by calculating the neighbors of test images in projected space. Gait recognition \cite{Xing2016} across various viewing angles was accomplished by first finding a unified feature space, and then a nearest neighbor classifier was used.

In addition to computer vision applications, many other areas also used CCA-based MVL approaches. Take phonetic recognition as example, it has been shown that the phonetic recognition accuracy can be improved by learning both acoustic and articulatory features with CCA \cite{Wang2016, Wang2015Unsupervised, Arora2013}. There are many other similar MVL problems, include handwritten digit recognition \cite{Elmadany2016, Sun2008, Shin2011, Sun2016}, biometric structure prediction \cite{Luo2015}, text categorization \cite{Sun2008}, advertisement classification \cite{Luo2015}, and so on.

Generally, in a variety of multi-view recognition tasks, the CCA-based MVL approaches are first used to learn different new feature representations for each view, namely canonical variables. Once the canonical variables are obtained, there are two representative approaches for the subsequent learning. Take classification as example:
\begin{enumerate}
\item Directly put concatenated canonical variables of different views $[Z_{x}; Z_{y}]$ into SVM, KNN or other classifiers \cite{Luo2015,Shin2011,Shan2008}.
\item Take canonical variables of the first view $Z_x$ as input to a classifier \cite{Wang2016, Sun2008, Wang2015DCCAE}. There are two cases to choose this approach. The first is that in a real-world situation, only one view data is provided in the test. The second is to use target as the second view when applying the CCA-based approaches, so the prediction of test instance is accomplished by calculating the nearest neighbors in the first projected view.
\end{enumerate}

\subsection{Cross-modal Retrieval and Classification}

Due to the rapid growth of multimedia data on the Internet, cross-media retrieval has become a research hotspot in the fields of
artificial intelligence, multimedia information retrieval and computer vision. The main challenge of cross-media retrieval is the ``heterogeneity gap", which means that the data from different media types is inconsistent. This gap makes it diffcult to measure the cross-media similarity of instances \cite{Chi2017}.For example, a document containing images and text, and our goal is to retrieve text articles in response to query images and vice-versa. One of the traditional solutions is to find a common representation for different views with CCA \cite{Rasiwasia2010, Peng2017, Wu2006}. Firstly, taking images and text as two different views, the $k$-dimensional canonical variables of each instances are obtained with CCA-based MVL methods. Consider that all canonical variables from different views as coordinates in a common correlated space, as is shown in \ref{fig:CCA}. When given a query image, after mapping onto the common correlated space with canonical vectors,  the most closely matches can be obtained by computing its nearest neighbors in the range of all text canonical variables.

Cross-model classification \cite{Chandar2016} has the same data distribution assumption that the paired canonical variables of two different views are close to each other in the common correlated space. Therefore, a classifier trained on the first view can perform well in the second view.

\subsection{Multi-view Embedding}

One of the important applications of CCA-based MVL approaches is word embedding in many natural language processing tasks \cite{Dhillon2011, Lu2015, Faruqui2014}. Faruqui and Dyer \cite{Faruqui2014} proposed to learn words embeddings by adding multilingual context with DCCA, and found that the resulting embeddings could improve the performance on word similarity tasks. Benton \emph{et al.} \cite{Benton2016} learned multi-view embeddings of social media users to provided a solution for friend recommendation with GCCA \cite{Benton2017}.

In generally, multi-view embedding have two benefits. The first is to capture information from multiple views to obtain a representation with a better measure of similarity. The second is to learn a low-dimensional representation for high-dimensional multi-view data, which is critical to reducing the complexity of algorithms.

\subsection{Summary}
This section has introduced three main application of CCA-based MVL approaches, including multi-view recognition tasks, cross-modal retrieval/classificaion, and multi-view embedding. In addition to the applications described above, there are some other applications, including phonetic transcription \cite{Benton2017} and super-resolution images reconstruction \cite{An2014, Huang2010}, and so on.

\section{Conclusion} \label{sec:Conclusion}

CCA is widely used in MVL. However, traditional CCA has some limitations: 1) it can only handle two views; 2)  it can only optimize the linear correlation between different views; 2) it is unsupervised, and hence cannot make use of the label information. Many nonlinear, supervised, or generalized CCA extensions have been proposed for remedy. This paper has presented a survey of representative CCA-based MVL approaches.



\end{document}